\documentclass[runningheads]{llncs}
\usepackage{marvosym}
\usepackage[T1]{fontenc}
\usepackage{graphicx}
\usepackage{amssymb} 
\usepackage{pifont} 
\usepackage{amsmath} 
\usepackage{booktabs} 
\usepackage{multirow} 
\usepackage{arydshln} 

\usepackage{hyperref}
\hypersetup{
        hypertex=true,
        colorlinks=true,
        linkcolor=blue,
        anchorcolor=blue,
        citecolor=blue
}

\begin{document}
\title{EIFNet: Leveraging Event-Image Fusion for Robust Semantic Segmentation}
\titlerunning{EIFNet: Leveraging Event-Image Fusion for Robust Semantic Segmentation }
%
\author{Zhijiang Li\textsuperscript{\Letter}\inst{1} \and
Haoran He\inst{1} \and
Guoshun Nan\inst{1}}
\authorrunning{Z. Li et al.}
%
\institute{Beijing University of Posts and Telecommunications, Beijing 100876, China\\
\email{20030902@bupt.edu.cn, hehaoran666@bupt.edu.cn, nanguo2021@bupt.edu.cn}}

\maketitle              
\begin{abstract}

Event-based semantic segmentation explores the potential of event cameras, which offer high dynamic range and fine temporal resolution, to achieve robust scene understanding in challenging environments. Despite these advantages, the task remains difficult due to two main challenges: extracting reliable features from sparse and noisy event streams, and effectively fusing them with dense, semantically rich image data that differ in structure and representation. To address these issues, we propose EIFNet, a multi-modal fusion network that combines the strengths of both event and frame-based inputs. The network includes an Adaptive Event Feature Refinement Module (AEFRM), which improves event representations through multi-scale activity modeling and spatial attention. In addition, we introduce a Modality-Adaptive Recalibration Module (MARM) and a Multi-Head Attention Gated Fusion Module (MGFM), which align and integrate features across modalities using attention mechanisms and gated fusion strategies. Experiments on DDD17-Semantic and DSEC-Semantic datasets show that EIFNet achieves state-of-the-art performance, demonstrating its effectiveness in event-based semantic segmentation.

\end{abstract}

\section{Introduction}
Semantic segmentation plays a central role in visual understanding by predicting pixel-level semantic labels. It is a key component in applications such as autonomous driving, robotics, and scene interpretation \cite{r1}. While recent advancements in deep learning have pushed performance boundaries on standard benchmarks \cite{r2}, \cite{r3}, \cite{r4}, most approaches rely heavily on high-quality RGB images. This reliance makes them vulnerable in challenging environments involving motion blur, low illumination, or high-speed dynamics, where conventional frame-based cameras often fail due to limited temporal resolution and dynamic range.

Event cameras offer a biologically inspired alternative, capturing per-pixel brightness changes with microsecond latency and wide dynamic range \cite{r5}. These sensors are well-suited for dynamic or low-light scenarios, where they provide complementary motion information to frame-based inputs. However, event data are inherently sparse, asynchronous, and noisy, which limits their standalone semantic expressiveness and poses difficulties for conventional segmentation networks.

Recent research has explored multi-modal fusion of events and images to leverage their complementary advantages \cite{r9}, \cite{biswas2024halsie}, \cite{r11}, \cite{r12}. Despite progress, two critical challenges remain unresolved:
(i) Extracting robust, noise-resilient semantic features from sparse and noisy event inputs.
(ii) Effectively aligning and fusing heterogeneous features from the event and image domains, given their differing structure, density, and semantic granularity.

To tackle the challenges posed by the sparse, asynchronous, and noisy nature of event data, existing methods often rely on fixed intermediate representations such as event histograms \cite{r8}, \cite{r7}, six-channel images \cite{r6}, or voxel grids \cite{r9}, \cite{biswas2024halsie}, \cite{r11}. These representations are typically constructed by simply accumulating event counts or timestamps, which limits their ability to distinguish noise from meaningful motion. As a result, scene modeling becomes less reliable, and critical low-level features like edges and corners are easily lost in highly dynamic or low-light conditions. To address these limitations, we propose a novel representation approach named Adaptive Event Feature Refinement Module (AEFRM), which combines multi-scale activity modeling with channel attention mechanisms to dynamically capture the spatiotemporal structure of the event stream. By integrating local event density and global channel dependencies, it generates pixel-wise attention masks that adaptively enhance high-confidence motion regions while suppressing background noise, enabling more precise and robust event representations in complex scenes \cite{r12}.

In addressing the challenge of multi-modal feature fusion, existing methods typically rely on fixed-scale convolutions or shallow cross-attention \cite{r13}, \cite{r14}, resulting in insufficient depth of cross-modal interaction and difficulty in fully aligning sparse event semantics with dense image features. To address this, we propose a noval dual-stage fusion strategy with the Modality-Adaptive Recalibration Module (MARM) and the Multi-Head attention Gated Fusion Module (MGFM). MARM performs channel-space recalibration, optimizing semantic consistency for both event and image features while preventing noise propagation between modalities. MGFM introduces differential attention and cross-attention mechanisms to model global cross-modal semantic relationships and performs fine-grained feature fusion via dual-domain gating weights.

Based on these modules, we construct an \textbf{E}vent-\textbf{I}mage \textbf{F}usion \textbf{N}etwork (\textbf{EIFNet}), a dual-encoder architecture that progressively integrates multi-modal information through a ``representation-recalibration-fusion'' strategy. Experimental results demonstrate that EIFNet achieves substantial performance gains over existing methods, while comprehensive ablation studies systematically validate the effectiveness of each module and their synergistic interactions within the network. Compared to prior approaches, EIFNet offers distinct advantages in multi-scale noise suppression, deep cross-modal interaction, and dynamic gated fusion, effectively addressing challenges like noise, feature density differences, and semantic conflicts. 

The main contributions of this paper are summarized as follows:

\begin{itemize}
        \item We propose a new event representation approach (AEFRM) to enhance motion features and suppress noise through multi-scale modeling and spatial attention in dynamic scenes.
        \item We design a modality-adaptive recalibration module (MARM) to recalibrate event and image features via channel and spatial attention, effectively reducing noise and improving semantic consistency.
        \item We introduce a novel event-image fusion module (MGFM) to align and integrate event and image features.   It combines cross attention and differential attention mechanisms, leveraging adaptive gating for precise feature aggregation based on the distinct characteristics of each modality.
        \item Experimental results show that the proposed network (EIFNet) achieves outstanding performance for event-image based semantic segmentation on two benchmark datasets.
\end{itemize}

\section{Related Work}
\subsection{Event-Based Representation Learning}
Event cameras have received increasing attention in recent years due to their high dynamic range, microsecond-level temporal resolution, and ultra-low latency \cite{r15}, \cite{r16}, \cite{r17}. Unlike conventional cameras, event cameras asynchronously record brightness changes at each pixel and output sparse event streams, providing unique advantages for tasks such as motion detection and edge-aware perception. However, the sparse and high-frequency nature of event data poses challenges in achieving stable and noise-robust feature representations.

Early approaches relied on handcrafted representations of event data, such as event histograms, time surfaces, and polarity maps, which convert the event stream into 2D dense tensors compatible with standard convolutional networks \cite{r8}, \cite{r7}, \cite{r6}. Later methods introduced voxel grid-based spatio-temporal encoding by discretizing the event stream into fixed-length pseudo-image sequences over time \cite{r11}. While these techniques improve compatibility with deep models, they often struggle to suppress noise and extract meaningful information in dynamic scenes. To address these issues, recent studies have incorporated attention mechanisms and learned representations to improve the quality of event features. For example, EISNet introduced an activity-based AEIM module to enhance salient regions based on event density \cite{r12}. However, its fusion remains largely constrained to fixed-scale convolutions and limited interaction depth.

In this work, we propose AEFRM, which combines multi-scale activity modeling and spatial attention to dynamically estimate the confidence of each channel region. A pixel-wise attention mask is generated to selectively enhance informative structures while suppressing background noise. This approach produces a more stable and robust event representation that lays a strong foundation for subsequent multi-modal fusion.

\subsection{Multi-Modal Semantic Segmentation}
Multi-modal semantic segmentation seeks to enhance robustness under complex conditions by fusing complementary information from various sensors such as RGB images, depth maps, thermal images, or event data. Existing multi-modal fusion methods can be generally classified into two categories: unidirectional enhancement and bidirectional interaction strategies.

Unidirectional enhancement approaches typically treat one modality as primary (e.g., RGB) and incorporate auxiliary modalities (e.g., events) via feature-level attention or weighted summation \cite{r2}, \cite{r3}, \cite{r4}. While simple, these methods often underutilize the auxiliary modality and may lead to modality dominance \cite{r18}, where critical information from the secondary modality is suppressed.

To enable stronger cross-modal interaction, bidirectional fusion methods have gained popularity. Representative works such as CMX \cite{r13} and CMNeXt \cite{r14} use intermediate fusion and cross-attention modules to enable information exchange between modalities \cite{r19}, \cite{r20}, \cite{r21}. However, most of these models are designed for dense modality pairs (e.g., RGB-D, RGB-T), and may not preserve the unique structure of sparse inputs like events. Some even treat event data as static pseudo-images, failing to exploit their temporal characteristics.

To address the significant differences in density and semantics between events and images, we propose a two-stage fusion strategy: MARM performs dynamic recalibration for each modality across both channel and spatial dimensions, effectively suppressing unreliable regions while enhancing discriminative features. MGFM leverages cross-attention and differential attention to establish global interaction pathways between modalities \cite{r12}, \cite{r22}. A dual-domain gating mechanism further generates spatial and channel attention maps to guide accurate and complementary feature fusion. This fusion mechanism both strengthens modality-specific representations and forms efficient, structure-preserving cross-modal interaction pathways, maximizing event - image fusion potential.

\section{Methodology}\label{sec:methodology}
Section~\ref{sec:overall_architecture} outlines the overall architecture of the proposed EIFNet model for event-image semantic segmentation. Section~\ref{sec:aefrm} explains AEFRM for enhancing event features. Section~\ref{sec:marm} introduces MARM for modality-specific recalibration. Section~\ref{sec:mgfm} details MGFM for bidirectional feature fusion.

\subsection{Overall Architecture}\label{sec:overall_architecture}
\begin{figure}[t]
\centering
\includegraphics[width=\textwidth]{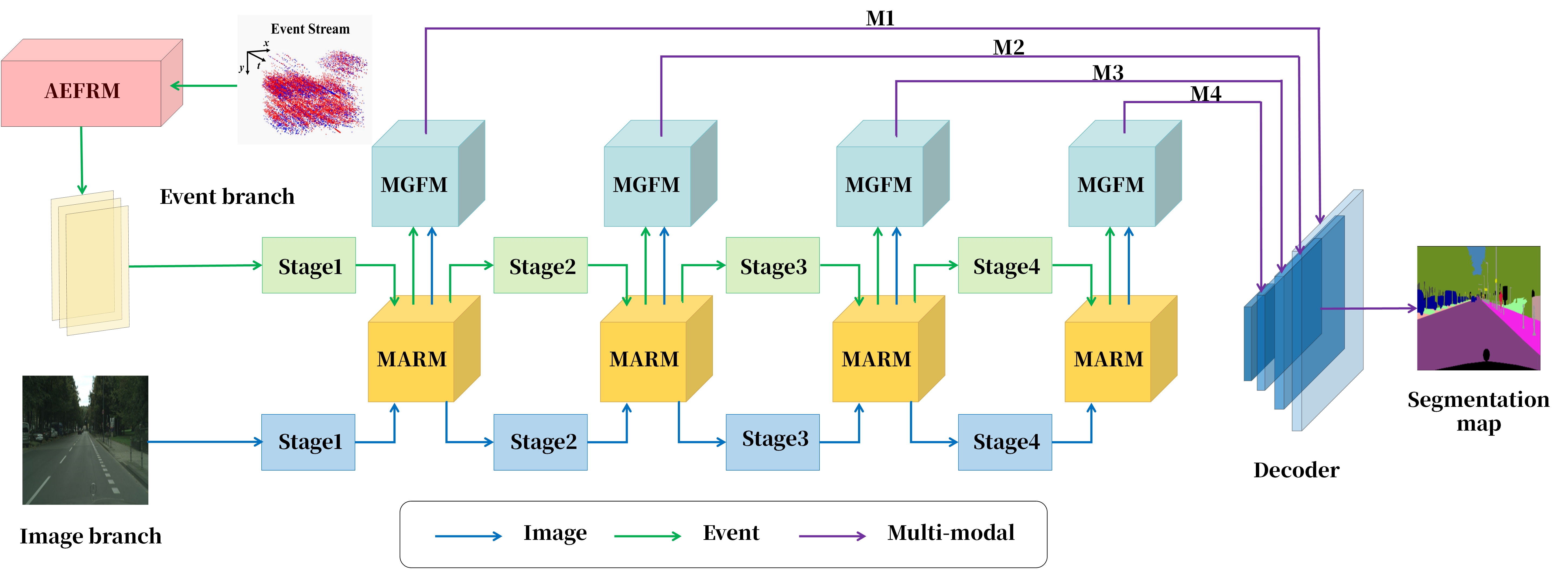}
\caption{Overall architecture of EIFNet. The network consists of an AEFRM module for event enhancement, dual-branch Transformer encoders, four stages of recalibration (MARM) and gated fusion (MGFM), and a lightweight decoder for final segmentation.}
\label{fig:overall_architecture}
\vspace{-10pt}
\end{figure}

As shown in Fig.~\ref{fig:overall_architecture}, the overall structure of EIFNet is designed to fully exploit the complementary properties of event and image modalities, particularly in challenging conditions such as low lighting and rapid motion. The backbone consists of dual encoders for each modality, a multi-stage fusion pipeline, and a Transformer-based decoder for final prediction.

This network takes synchronized event and image frames as input. AEFRM processes event data by multi - scale modeling and spatial attention to suppress noise and enhance structural info, producing robust event features. Then, these features and raw image features are fed into dual - branch Transformer encoders with MiT backbones of different depths. The event branch captures motion boundaries and sparse activations, and the image branch extracts dense texture and color semantics.

At each encoder stage, the outputs from both branches are passed through MARM, which performs channel-wise and spatial-wise recalibration to enhance semantic relevance and suppress redundant features. The recalibrated features are then passed to MGFM, where cross-attention and differential attention mechanisms model global context exchange between modalities. A gated fusion unit combines the features adaptively using learned attention weights.

The fused multi-scale features from all stages $\{M_1, M_2, M_3, M_4\}$ are aggregated and passed into a lightweight Transformer decoder to restore spatial resolution and predict pixel-wise segmentation maps.

\begin{figure}[t]
\centering
\includegraphics[width=\textwidth]{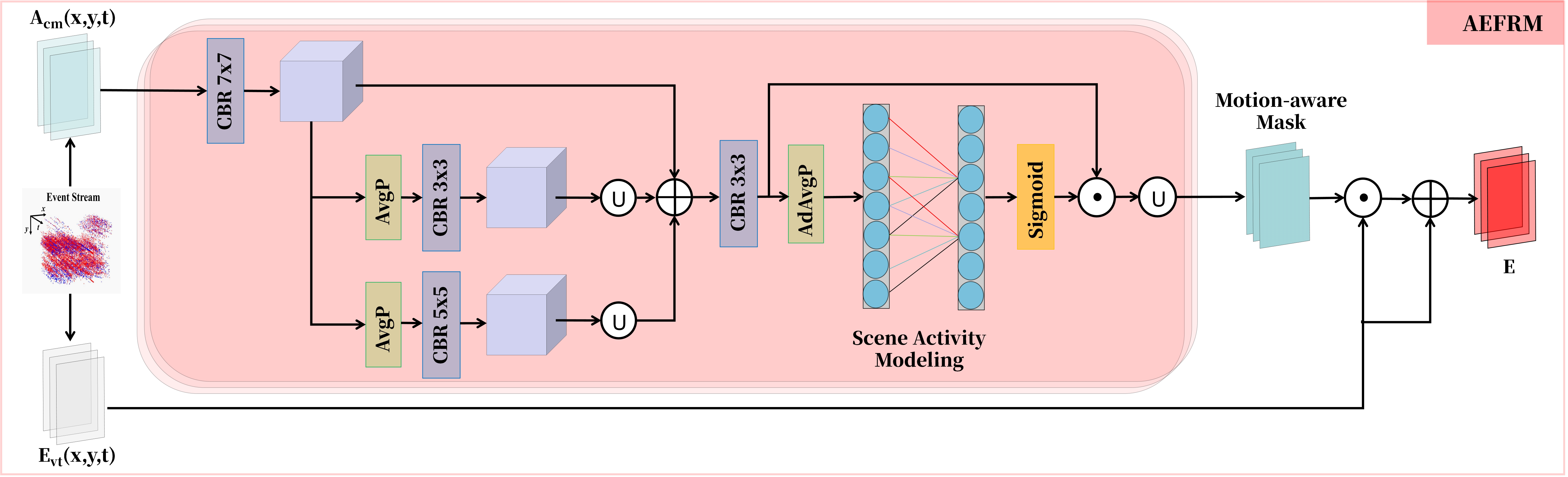}
\caption{Structure of AEFRM. The module enhances event features using multi-scale pooling and channel-wise attention. Outputs are adaptively fused with the input to suppress noise and highlight salient motion regions.}
\label{fig:AEFRM_structure}
\end{figure}

\subsection{Adaptive Event Feature Refinement Module (AEFRM)}
\label{sec:aefrm}

Event data carries sparse and asynchronous temporal information that highlights motion boundaries but also introduces noise. Direct projection into image-like forms may lose key temporal cues. To address this, AEFRM enhances salient structures and suppresses noise by combining multi-scale activity modeling with channel-wise attention.

We define the polarity-aware event projection \(E_{vt}\) and activity accumulation map \(A_{cm}\) as:
\begin{equation}
E_{vt}(x,y,t) = \sum_{i=1}^M p_i \delta(x - x_i, y - y_i) k(t - t_i^*), 
\end{equation}

\begin{equation}
A_{cm}(x,y,t) = \sum_{i=1}^M \delta(x - x_i, y - y_i) k(t - t_i^*),
\end{equation}
where $p_i \in \{-1,+1\}$ is event polarity, $(x_i,y_i,t_i^*)$ are spatio-temporal coordinates, and $k(z) = \max(0,1-|z|)$. The activity map $A_{cm} \in \mathbb{R}^{B \times C \times H \times W}$ is then reshaped to a pseudo-image tensor and processed by three parallel branches: a $7 \times 7$ convolution (stride 4) stem producing $F_s$, and two average pooling branches (with kernel sizes $3 \times 3$ and $5 \times 5$) followed by $3 \times 3$ convolutions producing $F_{p1}$ and $F_{p2}$. The fused feature map from the combined outputs is processed through a convolution block (CBR) of 3 × 3 convolution, batch normalization, and ReLU activation, resulting in the final fused feature map $F$, as shown in the equation below:
\begin{equation}
F = \text{CBR}\bigl(F_s + \text{Up}(F_{p1}) + \text{Up}(F_{p2})\bigr).
\end{equation}

To adaptively reweight the channel importance of the feature map \( F \in \mathbb{R}^{(B \cdot C) \times H \times W} \), we adopt a lightweight attention mechanism that combines global average pooling and a local 1 × 1 convolution with sigmoid activation. The attention weights are computed as:
\begin{equation}
W = \sigma\bigl(\text{Conv}_{1 \times 1}(\text{GAP}(F))\bigr),
\end{equation}
where \( \text{GAP} \) denotes global average pooling and \( \sigma(\cdot) \) is the sigmoid function. These weights are used to recalibrate the input feature and generate a spatial attention map in one step:
\begin{equation}
M = \text{Conv}_{1 \times 1}(F \otimes W),
\end{equation}
where \( \otimes \) denotes channel-wise multiplication. The resulting attention map \( M \) is upsampled and reshaped to match the resolution of the original input event tensor \( E_{vt} \in \mathbb{R}^{B \times C \times H \times W} \). Finally, the enhanced event representation is computed as:
\begin{equation}
E = E_{vt} \odot M + E_{vt},
\end{equation}
where \( \odot \) denotes element-wise multiplication. This operation highlights salient motion patterns while effectively suppressing background noise.

\begin{figure}[t]
\centering
\includegraphics[width=\textwidth]{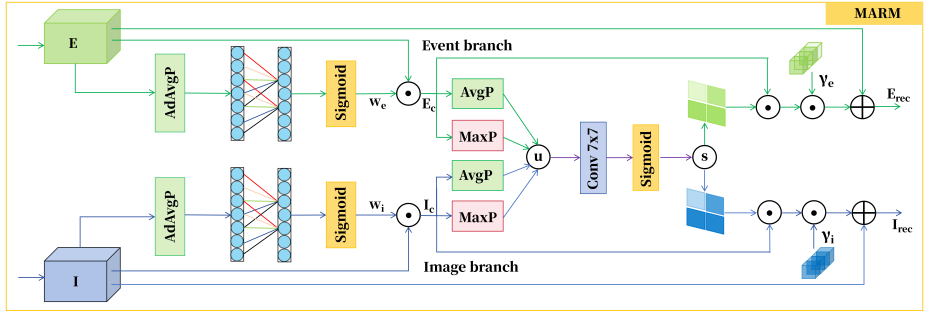}
\caption{Structure of MARM. The module performs channel-wise recalibration, followed by spatial attention via joint pooling and convolution. Both modalities are recalibrated independently to suppress noise and highlight informative regions.}
\label{fig:MARM_structure}
\end{figure}

\subsection{Modality-Adaptive Recalibration Module (MARM)}\label{sec:marm}
Event features are often sparse and locally triggered, while image features are dense and contain structured texture semantics. Due to the inherent differences in feature distribution, semantic granularity, and noise characteristics, direct fusion of these two modalities may lead to information conflicts and degraded consistency.

To mitigate this, we introduce MARM, which performs recalibration for each modality at the channel and spatial levels. This strategy enhances salient modality-specific regions while suppressing irrelevant or noisy features, preparing cleaner representations for subsequent cross-modal fusion. As illustrated in Fig.~\ref{fig:MARM_structure}, MARM consists of two main components: channel recalibration and spatial recalibration.

For channel recalibration, given an event feature map \(E \in \mathbb{R}^{B \times C_e \times H \times W}\) and an image feature map \(I \in \mathbb{R}^{B \times C_i \times H \times W}\), we first apply a channel attention mechanism to each modality separately. The mechanism uses a convolutional layer with a specific kernel size to capture global channel-wise dependencies:
\begin{equation}
w_e = \sigma(\text{Conv}_{1 \times 1}(\text{GAP}(E))), \quad w_i = \sigma(\text{Conv}_{1 \times 1}(\text{GAP}(I))).
\end{equation}
The resulting attention vectors \(w_e \in \mathbb{R}^{C_e}, w_i \in \mathbb{R}^{C_i}\) are broadcast-multiplied with the respective input features:
\begin{equation}
E_c = E \odot w_e, \quad I_c = I \odot w_i.
\end{equation}

For spatial recalibration, we further apply spatial attention to capture critical location-specific structures. To do this, we concatenate average and max-pooled features from both modalities:
\begin{equation}
S = \text{Concat}[\text{Avg}(E_c), \text{Max}(E_c), \text{Avg}(I_c), \text{Max}(I_c)] \in \mathbb{R}^{B \times 4 \times H \times W}.
\end{equation}
This tensor is passed through a \(7 \times 7\) convolution and sigmoid activation to generate spatial attention maps \(A_{sm}\):
\begin{equation}
A_{sm} = \sigma(\text{Conv}_{7 \times 7}(S)) \in \mathbb{R}^{B \times 2 \times H \times W}.
\end{equation}
The first channel corresponds to the event spatial attention map and the second to the image map. We apply these masks to recalibrate the respective features:
\begin{equation}
E_{rec} = E_c \odot A_{sm}^e \cdot \gamma_e + E, \quad I_{rec} = I_c \odot A_{sm}^i \cdot \gamma_i + I,
\end{equation}
where \(\gamma_e\) and \(\gamma_i\) are learnable scaling factors to balance the residual contributions. The outputs $E_{rec}$, $I_{rec}$ are the recalibrated event and image features passed on to the fusion stage. $A_{sm}^e$ and $A_{sm}^i$ are obtained by splitting $A_{sm}$ along the channel dimension.

\begin{figure}[t]
\centering
\includegraphics[width=\textwidth]{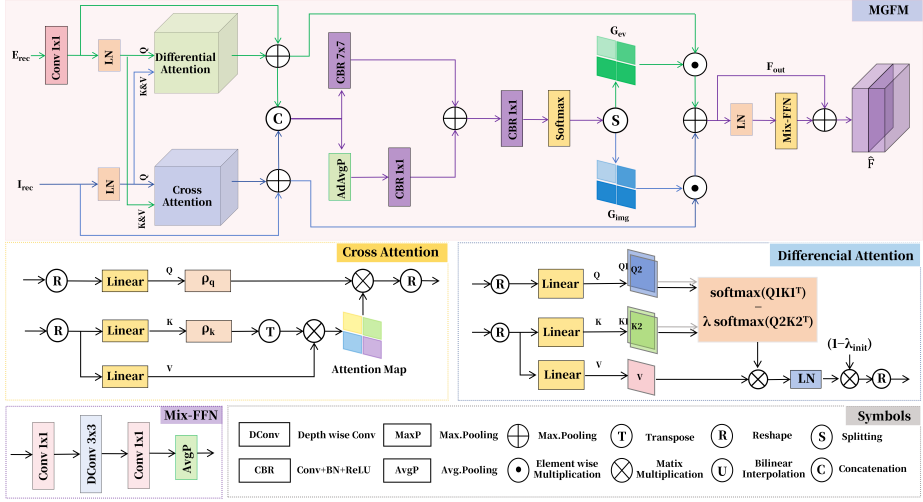}
\caption{Overview of MGFM and its attention components.  MGFM integrates event and image features using bidirectional attention and gated fusion.}
\label{fig:MGFM_overview}
\end{figure}

\subsection{Multi-head Attention Gated Fusion Module (MGFM)}\label{sec:mgfm}
Although the modality-specific features are independently recalibrated via MARM, significant differences remain between the dense semantic representations of images and the sparse, edge-focused features of events. To achieve deep, high-precision alignment and integration across modalities, we propose MGFM, which combines bidirectional cross-modal attention with a gated attention mechanism to extract complementary information and adaptively aggregate it. MGFM comprises three stages: modality interaction modeling, gated attention generation, and fusion execution.

For modality interaction modeling, given the recalibrated event and image features from MARM: $E_{rec} \in \mathbb{R}^{B \times C_e \times H \times W}$ and $I_{rec} \in \mathbb{R}^{B \times C_i \times H \times W}$, we apply Differential Attention and Efficient Cross-Attention to model bidirectional interactions:
\begin{equation}
E' = \text{DifferentialAttention}(E_{rec}, I_{rec}) + E_{rec},
\end{equation}
\begin{equation}
I' = \text{EfficientCrossAttention}(I_{rec}, E_{rec}) + I_{rec}.
\end{equation}

We specifically employ Differential Attention for event features because event data is sparse and prone to noise. Differential Attention effectively suppresses irrelevant noise by focusing on key feature differences, which is crucial for extracting meaningful motion patterns from event streams. Meanwhile, Efficient Cross-Attention is used for image features due to their dense semantic nature. It allows image features to incorporate complementary information from event data, enhancing semantic context. This dual approach not only refines modality-specific features but also constructs efficient cross-modal interaction pathways, maximizing the potential of event-image fusion for semantic segmentation.
These attention layers are implemented in a multi-head fashion, capturing semantic correspondences across modalities and enabling global information exchange.

In the spatial-Channel gated attention generation stage, the two attended features are concatenated to form a joint representation $F_{\text{fused}}$:
\begin{equation}
F_{\text{fused}} = \text{Concat}(E', I') \in \mathbb{R}^{B \times 2C \times H \times W},
\end{equation}
We then generate spatial and channel attention maps in parallel: Channel attention \(A_c\) is computed using global average pooling followed by a \(1 \times 1\) convolution, batch normalization, and ReLU activation. Spatial attention 
\(A_s\) is derived using a \(7 \times 7\) convolution, followed by a batch normalization, and ReLU activation:
\begin{equation}
A_c = \text{ReLU}(\text{BN}(\text{Conv}_{1 \times 1}(\text{GAP}(F_{\text{fused}})))) \in \mathbb{R}^{B \times 2 \times 1 \times 1},
\end{equation}
\begin{equation}
A_s = \text{ReLU}(\text{BN}(\text{Conv}_{7 \times 7}(F_{\text{fused}}))) \in \mathbb{R}^{B \times 2 \times H \times W},
\end{equation}
The sum of these two maps is passed through a \(1 \times 1\) convolution and a softmax function to obtain the final gating map \(G\):
\begin{equation}
G = \text{Softmax}(\text{Conv}_{1 \times 1}(A_c + A_s)) \in \mathbb{R}^{B \times 2 \times H \times W}.
\end{equation}

In the fusion execution and feature enhancement stage,
the two attention weights in \(G\) represent the relative contributions of the event and image features at each pixel. The final fused output \(F_{\text{out}}\) is computed as:
\begin{equation}
F_{\text{out}} = E' \odot G^e + I' \odot G^i,
\end{equation}
where $G^e$ and $G^i$ are obtained by splitting $G$ along the channel dimension. To further enhance the expressiveness of the fused features, we apply layer normalization and a feed-forward network (FFN):
\begin{equation}
\hat{F} = \text{FFN}(\text{LN}(F_{\text{out}})) + F_{\text{out}}.
\end{equation}
The features $\hat{F}$ obtained from each fusion stage are collected as $\{M_1, M_2, M_3, M_4\}$ and are aggregated before being passed into a lightweight Transformer decoder to restore spatial resolution and predict pixel-wise segmentation maps.

\begin{table}[t]
\caption{Semantic Segmentation Results of Different Methods in terms of mIoU and PA on the DDD17 and DSEC Datasets.}
\label{tab:comparison}
\centering
\resizebox{\textwidth}{!}{%
\begin{tabular}{lcccccc}
\toprule
\textbf{Method} & \textbf{Inference} & \textbf{Event Representation} & \multicolumn{2}{c}{\textbf{DDD17}} & \multicolumn{2}{c}{\textbf{DSEC}} \\
        &                    &                                & \textbf{mIoU (\%) ↑} & \textbf{PA (\%) ↑} & \textbf{mIoU (\%) ↑} & \textbf{PA (\%) ↑} \\
\midrule
SegFormer-B2    & Image              & -                              & 71.05 & 95.73 & 71.99 & 94.97 \\
SegNeXt-B       & Image              & -                              & 71.46 & 95.97 & 71.55 & 94.89 \\
\midrule
EV-SegNet       & Event              & 6-channel Image                & 54.81 & 89.76 & 51.76 & 88.61 \\
ESS             & Event              & Voxel Grid                     & 61.37 & 91.08 & 51.57 & 89.25 \\
\midrule
EDCNet-S2D      & Event + Image      & Voxel Grid                     & 61.99 & 93.80 & 56.75 & 92.39 \\
HALSIE          & Event + Image      & Voxel Grid                     & 60.66 & 92.50 & 52.43 & 89.01 \\
CMX             & Event + Image      & Voxel Grid                     & 71.88 & 95.64 & 72.42 & 95.07 \\
CMNeXt          & Event + Image      & Voxel Grid                     & 72.67 & 95.74 & 72.54 & 95.10 \\
EISNet          & Event + Image      & AET                            & 73.41 & 95.83 & 73.07 & 95.12 \\
Ours (EIFNet)  & Event + Image      & AEFRM                         & \textbf{76.56} & \textbf{96.18} & \textbf{74.64} & \textbf{95.61} \\
\bottomrule
\end{tabular}%
}
\end{table}

\section{EXPERIMENTS}
\subsection{Datasets}
We evaluate EIFNet on two widely used event-image semantic segmentation datasets: DDD17 Semantic \cite{r18} and DSEC-Semantic \cite{r7}. Both provide synchronized event-image pairs with pixel-level labels for diverse driving scenarios.

DDD17 Semantic: This dataset is derived from the DDD17 dataset and includes semantic labels for urban driving scenes under diverse lighting and motion conditions. It contains grayscale images, polarity event streams, and 6 semantic classes. We follow the standard protocol, using the provided training/testing split for evaluation.

DSEC-Semantic: Built on the DSEC dataset, DSEC-Semantic introduces pixel-level annotations for 11 semantic classes over day-time driving sequences. It includes RGB images and corresponding event data captured by color frame cameras and high-resolution monochrome event cameras. We follow the splitting strategy and data pre-processing method used in \cite{r7} to prepare the dataset for experiments.


\subsection{Implementation Details}

We implement EIFNet using PyTorch and train the model on RTX 4090. Training is performed using the AdamW optimizer with an initial learning rate of 0.0002 and a total of 60 epochs. The random seed is fixed to 1 to ensure reproducibility. We evaluate two datasets using the mean Intersection-over-Union (mIoU) and pixel-wise accuracy (PA).

For data preprocessing, we adopt random horizontal flipping and resizing as standard augmentations. Input images are center-cropped to \( 346 \times 260 \) for DDD17 and \( 640 \times 480 \) for DSEC-Semantic. Event streams are encoded using AEFRM, we use 3 temporal bins and a fixed integration window of 50 ms for encoding. 

The model employs dual MiT backbones: MiT-B0 for the event branch and MiT-B2 for the image branch \cite{r3}, both initialized from ImageNet-pretrained weights. Each modality is processed independently before entering the fusion modules. The training is conducted with a batch size of 16 and utilizes 8 data loader workers for efficient pipeline loading.

\subsection{Comparison with State-of-the-Art Methods}
We compare our proposed EIFNet with a range of state-of-the-art semantic segmentation methods on the DDD17 and DSEC-Semantic datasets. The evaluation covers three categories of models: (1) image-only baselines \cite{r3}, \cite{r4}, (2) event-only methods \cite{r7}, \cite{r6} and (3) event-image fusion approaches \cite{r9}, \cite{biswas2024halsie}, \cite{r11}, \cite{r12}.

As shown in Table~\ref{tab:comparison}, quantitative experiments demonstrate that EIFNet consistently outperforms existing models across both datasets in terms of mIoU and PA. Image-only methods like SegFormer-B2 achieve strong performance on DDD17 but underperform in challenging low-light or high-speed scenarios where event data is crucial. Event-only methods such as Event-Seg \cite{r6} and ESS \cite{r7} capture motion cues but lack semantic richness, resulting in lower segmentation quality. Fusion-based methods benefit from complementary modalities yet are limited by fixed voxel encodings or shallow interactions. EIFNet achieves new state-of-the-art mIoU of 76.56\% on DDD17 and 74.05\% on DSEC, with corresponding PA values of 96.18\% and 95.27\%, significantly outperforming all compared models.

\begin{figure}[t]
\centering
\includegraphics[width =\textwidth]{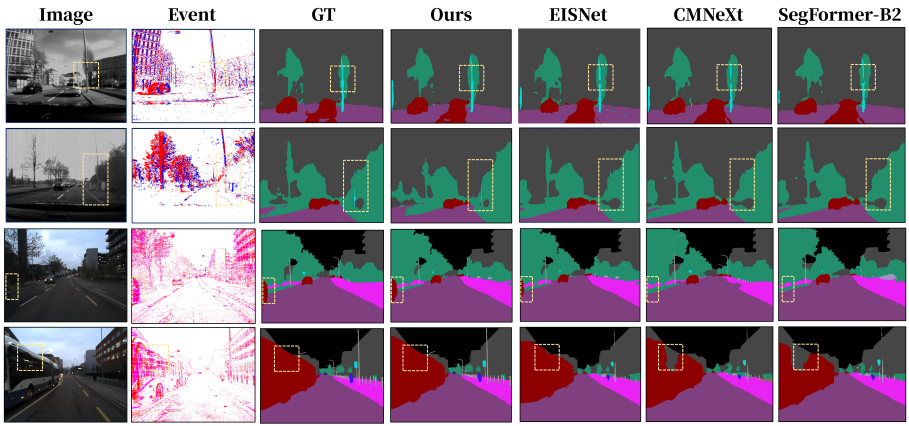}
\caption{Qualitative comparison of segmentation results on event-image inputs. From left to right: RGB image, event representation, ground truth (GT), predictions by our method (EIFNet), and by EISNet, CMNeXt, SegFormer-B2. EIFNet demonstrates clearer object boundaries and more accurate segmentation in challenging regions.}
\label{fig:qualitative_comparison}
\vspace{-10pt}
\end{figure}

Fig.~\ref{fig:qualitative_comparison} presents visual comparisons of our method with others across four challenging scenarios. Our method shows distinct advantages in generating accurate segmentation maps. For example, in row 2 where utility poles blend with complex backgrounds, our model achieves better prediction. Similarly, in row 4 with a fast-moving bus, our approach accurately segments them using event motion cues. These results intuitively demonstrate the effectiveness of our method in integrating complementary features for enhanced segmentation.

\begin{figure}[t]
\centering
\includegraphics[width=0.6\textwidth]{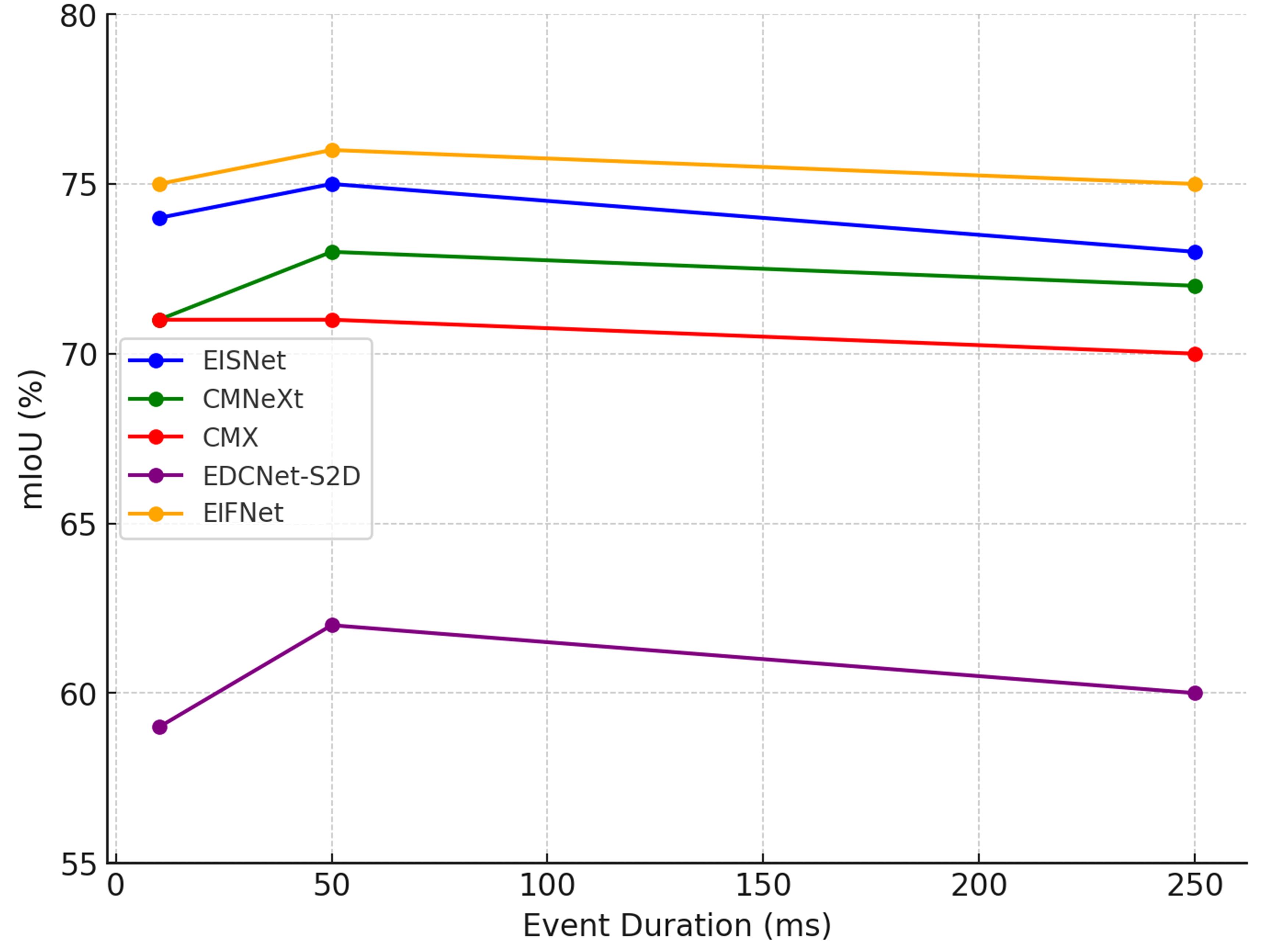}
\caption{Robustness performance of event-image based methods with different event duration on DDD17.}
\label{fig:mIoU_vs_Event_Duration}
\end{figure}

To evaluate the robustness of event-image based methods across varying time intervals, we conduct an experiment on DDD17. Models including EISNet, EDCNeXt-S2D, CMX, and CMNeXt are trained on 50 ms event data. During inference, they process image-event pairs with 10 ms and 250 ms event data. As shown in Fig.~\ref{fig:mIoU_vs_Event_Duration}, our model maintains superior and stable mIoU across intervals.

\begin{table}[t]
\caption{Ablation study of each proposed module on the DDD17 dataset.}
\label{tab:ablation_study}
\centering
\begin{tabular}{lccc|cc}
\toprule
\textbf{Method Variant} & \textbf{AEFRM} & \textbf{MARM} & \textbf{MGFM} & \textbf{mIoU (\%) ↑} & \textbf{PA (\%) ↑} \\
\midrule
Baseline + AEFRM        & \ding{51} & \ding{55} & \ding{55} & 74.69 & 96.03 \\
Baseline + MARM         & \ding{55} & \ding{51} & \ding{55} & 74.11 & 96.09 \\
Baseline + MGFM         & \ding{55} & \ding{55} & \ding{51} & 74.56 & 96.09 \\
Baseline + AEFRM + MARM & \ding{51} & \ding{51} & \ding{55} & 75.58 & 96.10 \\
Baseline + AEFRM + MGFM & \ding{51} & \ding{55} & \ding{51} & 74.94 & 95.68 \\
Baseline + MARM + MGFM  & \ding{55} & \ding{51} & \ding{51} & 76.36 & 96.08 \\
EIFNet (Full Model)    & \ding{51} & \ding{51} & \ding{51} & \textbf{76.55} & \textbf{96.19} \\
\bottomrule
\end{tabular}
\vspace{-10pt}
\end{table}

\subsection{Ablation Study}
To evaluate the effectiveness of each key component in EIFNet, we conduct a series of ablation studies on the DDD17 dataset, using EISNet as the baseline. We progressively integrate the proposed modules: AEFRM, MARM, and MGFM in the EISNet \cite{r12} architecture and assess performance gains in terms of mIoU and PA. As shown in Table~\ref{tab:ablation_study}, each proposed module contributes positively to the segmentation performance: Specifically, AEFRM adds +1.28\% in mIoU by enhancing sparse event features through activity-aware spatial attention. MARM contributes a +0.7\% improvement by suppressing modality-specific noise and highlighting informative regions. MGFM introduces a +1.15\% gain by modeling fine-grained cross-modal interactions and guided fusion. When all three modules are combined in the full EIFNet model, the mIoU further improves to 76.55\%, with PA reaching 96.19\%, demonstrating the synergistic benefits of the proposed components.

\begin{table}[t]
\caption{Model complexity and efficiency comparison.}
\label{tab:complexity_analysis}
\centering
\renewcommand{\arraystretch}{1.25}
\begin{tabular}{lcccc}
\toprule
\textbf{Method} & \textbf{Backbone} & \textbf{Params (M) ↓} & \textbf{MACs (G) ↓} & \textbf{mIoU (\%) ↑} \\
\midrule
EDCNet-S2D      & 2×ResNet-18   & \textbf{23.06} & \textbf{6.14} & 61.99 \\
CMX             & 2×MiT-B2      & 66.56          & 16.29         & 71.88 \\
\cdashline{1-5}
\multirow{2}{*}{CMNeXt} 
                                & E : PPX       & \multirow{2}{*}{58.68} & \multirow{2}{*}{16.32} & \multirow{2}{*}{72.67} \\
                                & I : MiT-B2    &                      &                      &                      \\
\cdashline{1-5}
\multirow{2}{*}{EISNet} 
                                & E : MiT-B0    & \multirow{2}{*}{34.39} & \multirow{2}{*}{17.30} & \multirow{2}{*}{73.41} \\
                                & I : MiT-B2    &                      &                      &                      \\
\cdashline{1-5}
\multirow{2}{*}{Ours (EIFNet)} 
                                & E : MiT-B0    & \multirow{2}{*}{35.48} & \multirow{2}{*}{17.89} & \multirow{2}{*}{\textbf{76.56}} \\
                                & I : MiT-B2    &                      &                      &                      \\
\bottomrule
\end{tabular}
\vspace{-10pt}
\end{table}

\subsection{Complexity Analysis}
We evaluate the computational complexity and efficiency of EIFNet and compare it with baseline fusion methods. Table~\ref{tab:complexity_analysis} summarizes the number of parameters, floating-point operations (FLOPs), and inference speed (FPS) on RTX 3090 GPU with input resolution \(346 \times 260\). Despite integrating multi-stage attention and fusion mechanisms, EIFNet maintains comparable complexity to EISNet. The overall inference speed remains above 55 FPS, which satisfies real-time requirements for most robotics and autonomous driving applications. The results indicate that EIFNet achieves a favorable trade-off between segmentation performance and computational cost. This makes it suitable for deployment in time-sensitive multi-modal perception systems.

\section{Conclusion}
In this paper, we propose EIFNet, a novel event-image semantic segmentation framework that leverages multi-stage attention and recalibration mechanisms to exploit cross-modal complementarity. The architecture integrates three key modules: AEFRM for enhanced motion-aware event representations, MARM for emphasizing informative regions, and MGFM for fine-grained feature interaction. Extensive experiments on the DDD17 and DSEC-Semantic datasets show that EIFNet achieves state-of-the-art performance while maintaining efficiency, ablation study demonstrates the contribution of each module. Future work will focus on deploying EIFNet in real-time robotics and autonomous driving systems, and extending it to instance-level and panoptic segmentation with event-aware cues.

%
%
%
%
%
\bibliographystyle{splncs04}
\bibliography{samplepaper}

\end{document}